# Explore the difficulty of words and its influential attributes based on the Wordle game

Beibei Liu*
*Beijing Institute of Technology*
Beijing,China
fcdlbb@126.com

Yuanfang Zhang*
*Beijing Institute of Technology*
Beijing,China
1120200746@bit.edu.cn

Shiyu Zhang*
*Beijing Institute of Technology*
Beijing,China
1120200721@bit.edu.cn

*Abstract*—We adopt the distribution and expectation of guessing times in game Wordle as metrics to predict the difficulty of words and explore their influence factors. In order to predict the difficulty distribution, we use Monte Carlo to simulate the guessing process of players and then narrow the gap between raw and actual distribution of guessing times for each word with Markov which generates the associativity of words. Afterwards, we take advantage of lasso regression to predict the deviation of guessing times expectation and quadratic programming to obtain the correction of the original distribution.

To predict the difficulty levels, we first use hierarchical clustering to classify the difficulty levels based on the expectation of guessing times. Afterwards we downscale the variables of lexical attributes based on factor analysis. Significant factors include the number of neighboring words, letter similarity, sub-string similarity, and word frequency. Finally, we build the relationship between lexical attributes and difficulty levels through ordered logistic regression.

*Keywords—assessment of vocabulary difficulty, word attributes, Monte Carlo, Markov, factor analysis, hierarchical clustering*

## I. INTRODUCTION

Wordle is a popular online word puzzle game. Its goal is to guess a hidden five letter word in no more than six guesses. After each guess, the computers reveals whether each letter is absent from the word (gray), present in the word but in a different position (yellow), or present in the current position (green). Wordle is unique in that it only allows one game to be played per day, and every player in the world plays to guess the same word each day [1].

Daily results of Wordle from January 7, 2022, to December 31, 2022, including the date, number of reported results on Twitter, word of each day and the times of guessing each word in one to more than six tries are provided by Twitter.

Since the average reported results is 90983, we hypothesize that the distribution and expectation of word guessing times in Wordle reflect the difficulty of words for the majority of people.

## II. LITERATURE REVIEWS

### A. Assess the difficulty of words

Being interested in factors that influence the difficulty of science words for students before and after learning, Gina N. [3] identified a set of word characteristics, including length, part of speech, frequency, morphological frequency, domain specificity, and concreteness. The variables measuring the difficulty of words were the pretest and posttest vocabulary scores of students. She adopted a series of stepwise regression analysis with polysemy, frequency and length as predictors of pretest and posttest scores. The result indicated that frequency and polysemy explained students' vocabulary growth scores over the course of instruction at two of three grade levels.

To further examine the predictive nature of word features that influence vocabulary difficulty, Hiebert [4] used a series of regression models to establish the relationship between student vocabulary performance, student age groups, and vocabulary attributes. His results indicated that word frequency and the corresponding age of students are worth consideration in the selection of words for instruction and for vocabulary assessments.

Brent's research [5] focused more on tests to evaluate vocabulary levels of Japanese students. She compared three common vocabulary test formats given to Japanese students, as measures of vocabulary difficulty. Brent added other variables of estimating word difficulty including transformed frequency, length of word, number of syllables and other orthographic features to correlation analysis, along with scores of three tests. Her result shows the orthographic features correlated highly with each other and the log of frequencies gives the best estimate of word difficulty.

Previous studies on assessing word difficulty have the following commonalities and limitations:

The study population is limited to groups of primary and secondary school students in specific countries and lacks population generalizability.

The indicators used to assess word difficulty are mostly the students' written vocabulary test scores, which are relatively homogeneous.

The selection of factors influencing word difficulty focused more on the nature of the words themselves and less on the connections between words.

Most of the research methods used correlation analysis of the variables and regression analysis of the influencing factors on difficulty. The large number of influencing factors and their multicollinearity make the results of regression analysis less accurate.

## B. The Wordle game

Since game wordle came up, some scholars have searched for optimal word guessing strategies. One of these strategies is based on the greedy algorithm. Martin B. [6] created two algorithms. The word guessing process is: The player selects the best word at each round in each game from the viable word sets. The player repeats the step until he/she reaches the point where the viable solution list has been pared down to one. Marti B. simulated the guessing process of all possible words in wordle and drew that the mean number of rounds to win was 3.7696.

The other optimal strategy to guess words is based on reinforcement learning. In Benton J.'s work [1], on the one hand, a metric was built for determining probability of a letter appearing as a green in a given letter position for the sequence of N words. On the other hand, Q learning was used to allow the player to make one of five types of guesses each round of the game based on rewards differentiated by the color of letters in the last word. A total win rate of 64.8% was achieved after 10000 trials using the best parameters.

Ivan Li [7] analyzed impacts of linguistic properties of words on player success. She used linear regression models to determine relationships between the average score that Twitter users scored daily and properties of the given word for the day. The result showed the number of orthographic neighbors a word has would negatively impact player's performance, while the frequency of word would positively impact plyer's performance.

To sum up, the previous studies on wordle mainly focused on finding the optimal strategy of the game, and there were few studies using the data of word guessing times to explore the factors affecting vocabulary difficulty.

## III. PREDICT THE DISTRIBUTION OF WORD DIFFICULTY

### A. Problem analysis

We need to develop a model that can predict the distribution of guessing times for given words. Our approach includes the following steps:

1) Simulate how people do the puzzle with Monte Carlo method and generated a rough draw of the distribution.

2) Analyze the relationship between the word and deviation that the raw distribution from the actual one, using metric based on Markov model.

3) Use Lasso regression to predict the deviation based on metrics that extract from words.

### B. Generate raw distribution with Monte Carlo Method

Monte Carlo method is a computational algorithm that rely on random sampling to obtain numerical results. The underlying concept is to use randomness to solve problems that might be deterministic in principle.

The strategy that the algorithm used in this task is as follows.

1) Set the solution word.

2) Set candidate word list according to the dictionary (includes 4,082 five-letter words).

3) Select a word from the list randomly.

4) Compare the selected word and the solution word and note letter restrictions.

5) Remove all the words that violate the restrictions in step 4 from the list.

6) Repeat step 2 to 5 until the answer is correct, count number of tries.

We ran the algorithm on all the 359 words in the data sheet and it was repeated 10,000 times on each word. With the number of tries returned by the function, we acquired the distribution of each word. The predicted distribution of solution word w is noted as $D'_w$, while the actual distribution is noted as $D_w$.

Surprisingly, the predicted distributions had errors that much lower than our expectation. Even some distributions that were against our common sense were correctly predicted.

For instance, intuitively, the more frequently the solution word of the puzzle is used, the easier players may solve the puzzle. As the frequently used words are more likely to be come up with compared to other words that the hints from the game lead to. However, the word *watch*, which is a common word, has a significantly higher expectation of number of tries as well as failure rate, which was correctly predicted by the algorithm.

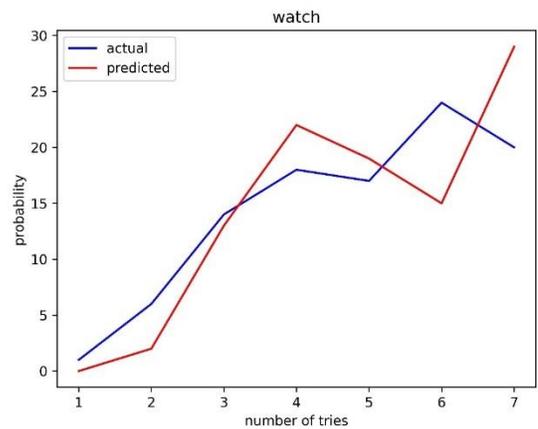

Fig.1. Comparison between actual distribution and predicted distribution of word *watch*.

Considering the strategy of the algorithm, this anomaly enlightened us that the confusability of a word is more decisive than the frequency of utility of it in the result of a game, i.e. if a solution word shares a similar structure or combination of letters with many other words, though players have played a few rounds and acquired hints, it's still hard for them to locate the answer from a vast set of probable words, even if the word is common. In contrast, if a solution word has an *exclusive* spelling, players won't be confused and can solve the puzzle swiftly, even if the word is rarely seen. So that, the result of word *watch* could be explained.

Although the algorithm worked well on some words, there are many other words have deviations between the predicted distributions and actual ones. Back to the algorithm, we believe the random selection of words is the main reason of the existence of deviation. Because when a real person selects a word from a set of words that are consistent with previous hint, the probabilities that which word will be come up with first are variant, which depends on his/her language habits and the word itself. Metrics that evaluate how easily a word w can be come up with is noted as $S_w$. The deviation of $D_w$ and $D'_w$ can be quantified as difference of expectation, noted as $E_{\Delta w}$.

$$E_{\Delta w} = \mathrm{E}(D'_w) - \mathrm{E}(D_w) \qquad (1)$$

We assume that $E_{\Delta w}$ is positively correlated with $S_w$. Because the higher the $S_w$, players are more likely to come up with solution w as the next answer which leads to a smaller $\mathrm{E}(D_w)$, while algorithm won't, so the bigger $E_{\Delta w}$ will be. Although confusability may also affect $D_w$ and $D'_w$, both players and the algorithm have taken it into account, so it's influence on $E_{\Delta w}$ can be ignored.

### C. Calculate associativity of words based on Markov Model

To prove the assumption above, we need to find out what $S_w$ consists of. We focused on how people come up with words.

To simplify this process, we regard the process that a person come up with a five-letter English word letter by letter as a Markov process [8]. A Markov process is a random process $\{X(t), t \in T\}$ that a sequence of events or states in which the probability of moving from one state to another depends only on the current state and not on any of the previous states. It follows the equation:

$$P\{X(t) = x | X(t_n) = x_n, \dots, X(t_1) = x_1\}$$
$$= P\{X(t) = x | X(t_n) = x_n\} \qquad (2)$$

In this problem, we build two matrixes $Mat_f$ and $Mat_t$ sized $1 \times 26$ and $26 \times 26$ respectively. The $i$-th element in $Mat_f$ represents the probability that the first letter that be come up with is the $i$-th letter in alphabet.

$$Mat_f(i) = P\{X(t_1) = l_i\} \qquad (3)$$

where $l_i$ represents the $i$-th letter in alphabet, $t_1$ represents the first letter and $X$ represents the process that a person comes up with a word letter by letter.

The element located at $i$-th row and $j$-th column in $Mat_t$ represents the probability of the next letter being the $j$-th letter, given that the previous letter is the $i$-th letter.

$$Mat_t(i, j) = P\{X(t_n) = l_j | X(t_{n-1}) = l_i\} \qquad (4)$$

where $n \leq 5$. The matrixes were built upon a dictionary that contains 4,082 five-letter words.

The probability of a word being generated through the process can be calculated as:

$$A(word) = Mat_f(word_1) \prod_{i=2}^{5} \mathrm{Mat_t}(word_{i-1}, word_i) \quad (5)$$

where $word_i$ represents the index of the $i$-th letter in the word in alphabet.

Here we called the product of the equation the Associativity of a word noted as $A_w$, which reflects how easily that a word can be come up with.

We calculated the associativity for all the solution words and made a logarithmic transformation because some of the values were extremely small. The relationship between word associativity and the deviation of expectation is presented below:

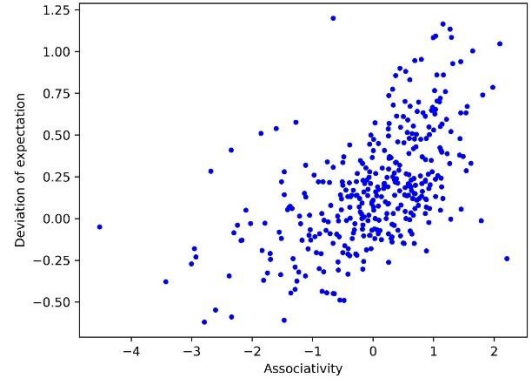

Fig.2. Scatter plot of the relationship between $E_{\Delta w}$ and $A$.

The correlation coefficient between the two variables is 0.556, which indicates that the easier it is to come up with a word, the bigger the $E_{\Delta w}$. This conclusion supports the validity of the assumption above.

### D. Use ease of thought to predict the deviation of average guessing times through Lasso Regression

Now that we had proved the relationship between $A_w$ and $E_{\Delta w}$, $A_w$ could be a component of $S_w$. In order to provide a more comprehensive measurement $S_w$. We chose word frequency noted $F_w$ as another metric of $S_w$. Because intuitively, the more often a word appears, the more familiar we become with it. Besides, these two metrics are irrelevant.

We used Lasso Regression to predict $E_{\Delta w}$ noted the prediction result as $E'_{\Delta w}$, which is a type of linear regression that includes a penalty term in the objective function to avoid overfitting. Another advantage of it is that it can perform feature selection.

We treated $A_w$ and $F_w$ as the independent variables and $E_{\Delta w}$ as the dependent variable and used the whole dataset as training set, the penalty parameter $\alpha = 0.01$. The mean squared error of the regression result is 0.069.

### E. Correct the ease of thought based on predicted deviation of average guessing times

Now we had had $E'_{\Delta w}$ from previous steps, we need to correct $D'_w$ by predicting the deviation that $D'_w$ from $D_w$ based on it. Here we noted the deviation as $D_{\Delta w}$, and the predicted $D_{\Delta w}$ as $D'_{\Delta w}$, which is of same size as $D_w$ and follow these constrains:

$$s.t. \begin{cases} \sum_{i=1}^{7} D'_{\Delta wi} = 0 \\ \sum_{i=1}^{7} i \cdot D'_{\Delta wi} = 100 E'_{\Delta w} \\ D'_{\Delta wi} \geq -D'_{wi} \quad (i = 1, 2, \ldots, 7) \end{cases} \quad (6)$$

We observed that the primary difference between $D_w$ and $D'_w$ lies in the offset of their peak positions, while the overall shape remains largely unchanged. Therefore, we attempted to make $D'_{\Delta w}$ at the corresponding positions positively correlated with $D'_w$, while minimizing the sum of squares of $D'_{\Delta w}$, which can be described as:

$$\min \sum_{i=1}^{7} c \cdot e^{-D'_{wi}} \cdot D'^2_{\Delta wi} \quad (7)$$

Thus, it can be solved as a quadratic programming problem. And the final prediction result can be calculated as $D'_w + D'_{\Delta w}$.

### F. Validation of the prediction model

We calculated the average mean squared error of the $D'_w$, $D'_w + D'_{\Delta w}$ and $D'_w + D'_{\Delta w}$ which used $E_{\Delta w}$ in constrains instead of $E'_{\Delta w}$ with $D_w$ for the whole dataset. The results are 31.93, 23.47 and 17.64 (probabilities range from 0 to 100) respectively, which proves the validity of the correction method.

### G. Prediction of word EERIE

TABLE I. PREDICTION OF THE WORD EERIE (AFTER ROUNDING)

| w | $D'_w$ | $A_w$ | $F_w$ | $E'_{\Delta w}$ | RESULT |
|---|--------|-------|-------|-----------------|--------|
| EERIE | [0, 1, 11, 33, 39, 14, 2] | 0.993 | -0.061 | 0.352 | [0, 0, 2, 32, 42, 18, 6] |

Based on the validation results, it can be reasonably guaranteed that the mean squared error of the outcomes is around **23**, which is a highly favorable result compared to other conventional prediction methods, such as neural networks.

## IV. PREDICT THE LEVEL OF WORD DIFFICULTY

In order to distinguish between difficulty levels and to properly assess the relationship between vocabulary and difficulty, we use a combination of three statistical methods: Hierarchical Clustering + Factor Analysis + Logistic Regression.

In order to distinguish difficulty levels and assess the relationship between vocabulary and difficulty properly, we use a combination of three statistical methods. First, we used Hierarchical Clustering for the dependent variables that can reflect the difficulty of vocabulary to classify each level of difficulty. Second, in order to evaluate the weight of vocabulary attribute, we artificially introduce relevant indicators of vocabulary attributes and use Factor Analysis to downscale it for the influencing factors. Finally, in order to establish the relationship between the influencing factors and the difficulty levels, we used Logistic Regression to analyze the degree of influence of each influence component on difficulty levels.

The result of the model shows four difficulty levels, and it is mainly decided by these factors: Number of Neighboring Words (NNW), Letter similarity (LS), String Similarity (SS), Word Frequency (WF).

### A. Generalizing vocabulary difficulty using hierarchical clustering

In this section, we generalize the vocabulary difficulty levels as reflected by the player data by means of hierarchical clustering.

We firstly map a table showing the relationship between silhouette score and number of clustering. Silhouette score is a way to evaluate the clustering effect, with the higher the value, the higher the accuracy of clustering.

It is obvious that categories of three or four can be the optimal choice for clustering. However, in order to make the sample more distinguishable, we tentatively classify the results with four categories. The assessment of clustering number will be reiterated in the third stage by Ordering Logistic Regression.

The conclusion in the previous question tells us that mathematical expectation and difficulty show a positive correlation. In that case, we need to evaluate the relationship between classification and expectation. If the expectation and classification results show a positive trend, it can be preliminary indicated that our classification method is feasible.

It is obvious that the two elements are positively correlated. The conclusion indirectly proves that 4-category hierarchical clustering has a certain amount of rationality and viability, shown as figure.3. below.

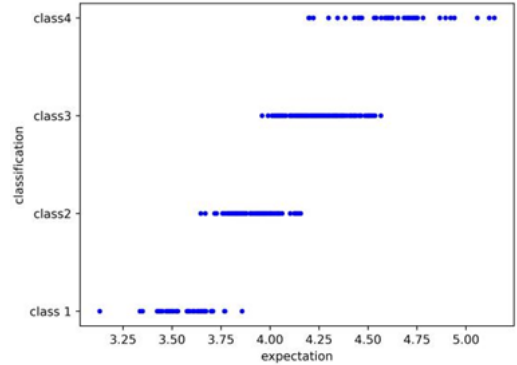

Fig.3. Preliminary evaluation of hierarchical clustering.

### B. Refining the main influential components of lexical attributes using Factor Analysis

In this section, we use factor analysis for the data to summarize the common indicators from these variables. Fig.4. is shown below.

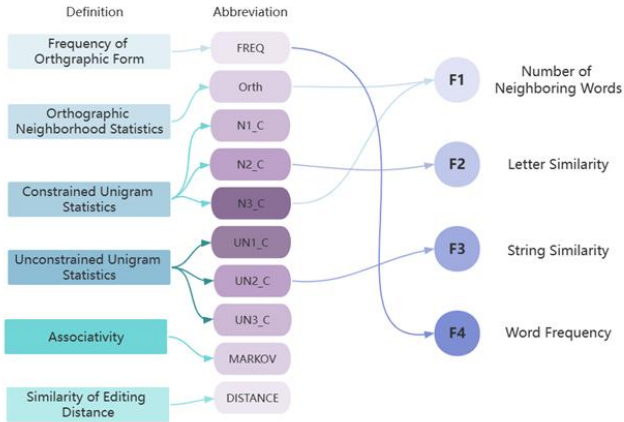

Fig.4. Flow chart for factor analysis.

We use factor analysis to downscale variables and simplifies the subsequent analysis process. Assume that there are random variables $X = (X_1, X_2, \ldots X_p)$ and the main factor is $F = (F_1, F_2, \ldots F_p)$. Assuming that the effects of all influential factors are linear, the above description can be expressed in mathematical terms as follows. $A = (\alpha_{ij})$ is called factor loading (Loading) [11].

$$X_i = \mu_i + \alpha_{i1}F_1 + \ldots + \alpha_{im}F_m + \varepsilon \qquad (8)$$

$$\begin{bmatrix} X_1 \\ X_2 \\ \ldots \\ X_p \end{bmatrix} = \begin{bmatrix} \mu_1 \\ \mu_2 \\ \ldots \\ \mu_p \end{bmatrix} + \begin{bmatrix} \alpha_{11}\alpha_{12}\cdots\alpha_{1m} \\ \alpha_{21}\alpha_{22}\cdots\alpha_{2m} \\ \ldots\cdots\cdots\cdots \\ \alpha_{p1}\alpha_{p2}\cdots\alpha_{pm} \end{bmatrix} * \begin{bmatrix} F_1 \\ F_2 \\ \ldots \\ F_p \end{bmatrix} + \begin{bmatrix} \varepsilon_1 \\ \varepsilon_2 \\ \ldots \\ \varepsilon_p \end{bmatrix} \qquad (9)$$

The explanations of initial variables we select are listed as follows.

FREQ: Frequency is a measure of how often a wordform is encountered in 1,000,000 presentations of text.

Orth: This is the number of orthographic neighbors that a string has. An orthographic neighbor is defined as a word of the same length that differs from the original string by only one letter.

Nx_C: This is a count of the number of wordforms that share the same constrained strings of x letters. A constrained string is defined as a specific letter in a specific position, in a specific length of word.

UNx_C: This is a count of the number of wordforms that share the same strings of x letters. An unconstrained string is defined as a specific letter within a word, regardless of its position, or the word length [9].

MARKOV: Associativity of words calculated in the last section.

DISTANCE: Similarity of editing distance between two words.

We estimated the factor loading matrix using factor analysis, and selected the main factors by the eigenvalues of the gravel plot and the factor contribution ratio. We extract N3_C and Orth from F1, N1_C from F2, UN2_C from F3 and FREQ from F4 as the maximum weighting factor among main factors.

Explanations of 4 main factors are listed as follows.

Number of Neighboring Words (NNW): Number of proximate words.

Letter similarity (LS): Number of a single letter repetitions in a word for certain position.

String Similarity (SS): Number of string repetitions in a word, regardless of position.

Word Frequency (WF): Mainly composed of Variables FREQ.

From the component matrix, we can come up with the principal component formulas:

$$\begin{cases} F1 = -0.003 FREQ + 0.196 Orth + 0.078 N1\_C + 0.15 N2\_C + 0.215 N3\_C - 0.07 UN1\_C \\ \qquad + 0.025 UN2\_C + 0.094 UN3\_C + 0.09 MARKOV - 0.073 DISTANCE \\ F2 = -0.026 FREQ + 0.135 Orth + 0.578 N1\_C + 0.299 N2\_C + 0.124 N3\_C + 0.426 UN1\_C \\ \qquad + 0.144 UN2\_C - 0.137 UN3\_C + 0.357 MARKOV - 0.56 DISTANCE \\ F3 = 0.035 FREQ + 0.062 Orth - 0.062 N1\_C + 0.25 N2\_C + 0.145 N3\_C + 0.46 UN1\_C \\ \qquad + 0.717 UN2\_C + 0.585 UN3\_C + 0.478 MARKOV - 0.146 DISTANCE \\ F4 = 1.013 FREQ + 0.024 Orth - 0.015 N1\_C - 0.038 N2\_C - 0.027 N3\_C + 0.038 UN1\_C \\ \qquad - 0.018 UN2\_C + 0.107 UN3\_C + 0.019 MARKOV - 0.2 DISTANCE \end{cases} \qquad (10)$$

We applied the scoring function to the mathematical model to obtain the four principal component values corresponding to all the words as the main influencing factors of vocabulary. These influencing factors are highly representative and sufficient to reflect the prominent attributes of the words.

### C. Analysis of factors' influence on the difficulty level using Ordered Logistic Regression

In this section, we will use ordered logistic regression to combine the influencing factors with difficulty levels. It is worth stating that influencing factors are F1-F4 based on factor analysis and difficulty levels are four categories based on hierarchy cluster.

By analyzing the ordered logistic regression results, we can learn the degree of influence of each influential factor on the ordered classification results, and list the weighting equations to measure the optimal classification model：

$$\begin{cases} y = \beta_0 + \beta_1 x_1 + \ldots + \beta_m x_m + \varepsilon \\ \varepsilon \sim N(0, \sigma^2) \end{cases} \qquad (11)$$

The classification task in this section is multivariate, so Ordered Logistic Regression is used.

The independent variables are the influencing factors F1-F4, and the dependent variables are the difficulty levels. And here is an intermediate variable that can be calculated throughout weighting functions:

$$y = 1.343 F1 + 0.823 F2 + 0.732 F3 + 0.687 F4 \qquad (12)$$

The table below shows the dependent variable classification thresholds. The difficulty levels are divided into 1 to 4 in total, with the higher the value, the higher the difficulty.

TABLE II.  CLASSIFICATION THRESHOLD OF DIFFERENT LEVELS

| Difficulty levels | 1 | 2 | 3 | 4 |
|---|---|---|---|---|
| Prediction value | $y \leq -2.20$ | $-2.20 < y \leq -0.32$ | $-0.32 < y \leq 2.00$ | $y > 2.00$ |

In the four classification tasks, the accuracy reaches 71.8% and the AUC=0.76 is in the range of [0.70, 0.85]. In summary,

the results of this ordered logistic regression are suitable and can be applied to the classification of vocabulary difficulty.

Then the question arises: As there are plenty of clustering models and regression models, how can we definitely confirm that we choose the most effective one? So we tested six sets of data from three different clustering approaches (K-Means, K-Shape, Hierarchy Cluster) and two different regression models (ordered logistic regression, logistic regression) permuted and combined, and drew radar plots as Fig.5. of the relevant evaluation metrics for each regression model.

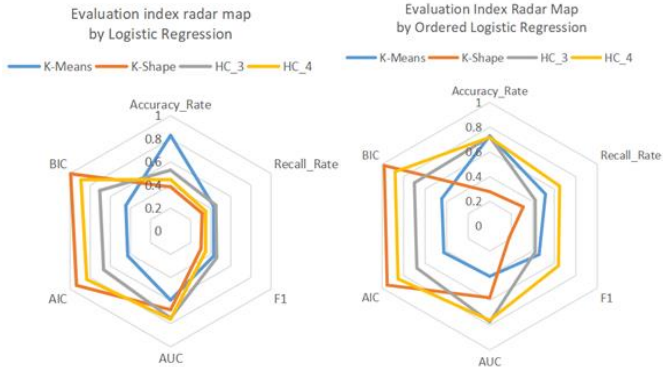

Fig.5. Radar maps of evaluation indexes.

The overall goal is to select the optimal combination of statistical combinations. Due to the character of word classification, our requirements for its accuracy are lower than those for the model's adequate(stability) styling . Therefore, when considering the optimal combination, priority should be given to the AIC, BIC, and AUC of the model, while taking into account the accuracy of the model. AIC, BIC and AUC are measures of fitness for statistical models: The higher the indicators are, the more adequate the model is.

For this reason, we analyzed each of the six combinations. On balance, we finally saved the model combination of "Four-category Hierarchical Clustering + Factor Analysis + Ordered Logistic Regression" to obtain a relatively high accuracy and stability.

### D. Defining the difficulty levels of EERIE

We follow the model above to find the normalized values, and thus the values of the principal factors.

Then, it is known that the predicted value.

$$y=1.343F1+0.823F2+0.732F3+0.687F4 = 3.236 \qquad (13)$$

Our calculation shows that value falls within region 4. Therefore we can conclude that EERIE should be the most difficult level.

## V. INTERESTING WORD ATTRIBUTES

In addition, we discover some interesting word attributes against our commonsense and previous studies.

Frequency is not the most influential factor of word difficulty.

Connections among words are more important than respective attributes of them.

Due to the unique guessing process of Wordle, the guessing time depends more on similarity of letters or strings.

## VI. DISCUSSIONS AND FUTURE EXTENSIONS

Our study has the following strengths, which makes up the shortcomings of previous studies.

Sufficient data source: Datasets are obtained from Twitter users of all ages around the world.

Novel indexes for assessment: We take the expectation and distribution of word guessing times in the wordle game as the measures of word difficulty. Moreover, we use variables of connections among words rather than their respective attributes to predict word difficulty, such as substring similarity and neighborhood words.

Innovative and mixed research methods: We construct a simulation-correction model to predict the distribution of word difficulty. Meanwhile, we build a feature extraction-dimensionality reduction-classification model to predict the levels of word difficulty.

Precision and accuracy: The MSE of the distribution of word difficulty is reduced to 0.23 from 0.32 after correction. The accuracy of levels of word difficulty reaches 0.72, proving that our metrics are reasonable and effective.

In the future, word difficulty levels and their influencing factors our study provided can be used for the word selection in educational word-guessing games as well as for the optimization of English vocabulary learning apps and teaching.

### ACKNOWLEDGMENT

We sincerely appreciate all the professors for helping us in our academic career and all the predecessors for providing such precious literature for us.